\pdfoutput=1

\documentclass[11pt]{article}

\usepackage[]{ACL2023}

\usepackage{times}
\usepackage{latexsym}

\usepackage[T1]{fontenc}

\usepackage[utf8]{inputenc}

\usepackage{microtype}

\usepackage{inconsolata}
\usepackage{multicol}
\usepackage{booktabs}
\usepackage{colortbl}
\usepackage[fleqn]{amsmath}
\usepackage{multirow}
\usepackage[normalem]{ulem}
\useunder{\uline}{\ul}{}
\usepackage{hyperref}
\usepackage{graphicx}
\usepackage{subfigure}
\usepackage{fixltx2e}
\usepackage{xcolor}
\usepackage{algorithm}
\usepackage{algorithmic}
\usepackage{setspace}
\usepackage{amsfonts}
\usepackage[most]{tcolorbox}

\newcommand{\ie}{\textit{i.e.}}
\newcommand{\eg}{\textit{e.g.}}
\newcommand{\ours}{\texttt{SPEED}}
\newcommand{\todo}[1]{\{\textcolor{blue}{\textbf{TODO}}\}}

\title{\textit{Little Giants}: Synthesizing High-Quality Embedding Data at Scale}

\author{Haonan Chen$^{1}$\thanks{$^{*}$Work done during Haonan’s internship at MSR Asia. Prof. Zhicheng Dou is the corresponding author.}, Liang Wang$^2$, Nan Yang$^2$, Yutao Zhu$^1$, \\ \textbf{Ziliang Zhao$^1$, Furu Wei$^2$, Zhicheng Dou$^{1}$} \\
        $^1$Gaoling School of Artificial Intelligence, Renmin University of China \\ 
        $^2$Microsoft Corporation \\ 
        \texttt{\{hnchen,dou\}@ruc.edu.cn} \\
        \texttt{\{wangliang,nanya,fuwei\}@microsoft.com} \\
}

\newtcolorbox[list inside=prompt]{prompt}[1][]{
    colbacktitle=black!60,
    coltitle=white,
    fontupper=\footnotesize,
    boxsep=5pt,
    left=0pt,
    right=-1pt,
    top=0pt,
    bottom=0pt,
    boxrule=1pt,
    #1,
}
\begin{document}
\maketitle

\begin{abstract}

Synthetic data generation has become an increasingly popular way of training models without the need for large, manually labeled datasets. For tasks like text embedding, synthetic data offers diverse and scalable training examples, significantly reducing the cost of human annotation. However, most current approaches rely heavily on proprietary models like GPT-4, which are expensive and inefficient for generating large-scale embedding data.
In this paper, we introduce \ours{}, a framework that aligns open-source small models (8B) to efficiently generate large-scale synthetic embedding data. Through supervised fine-tuning, preference optimization, and self-improvement, \ours{} enables small open-source models to produce high-quality data. 
Remarkably, \ours{} uses only less than 1/10 of the GPT API calls, outperforming the state-of-the-art embedding model E5$_\text{mistral}$ when both are trained solely on their synthetic data.
Using this efficient generator, we conduct a comprehensive study on how various factors within the alignment pipeline impact data quality and reveal the scaling law for synthetic embedding data.
Our codes and models are released in \url{https://github.com/haon-chen/SPEED}.

\end{abstract}
\section{Introduction}
Text embedding models encode natural language texts into latent vectors. 
They are widely used in downstream tasks such as classification, clustering, retrieval, and summarization.
Many researchers have trained general embedding models that can support various tasks~\cite{sbert,E5,bge}.
Most of these models require large-scale weakly-supervised data and high-quality labeled data for multi-stage training, which requires careful data curation and costly human effort.
Thanks to the powerful language modeling ability and vast knowledge of large language models (LLMs), some works attempt to utilize LLMs to generate synthetic data for training embedding models~\cite{InParsv2,E5mistral,Gecko}.

\begin{figure}[!t]
	\centering
	\includegraphics[width=1.0\linewidth]{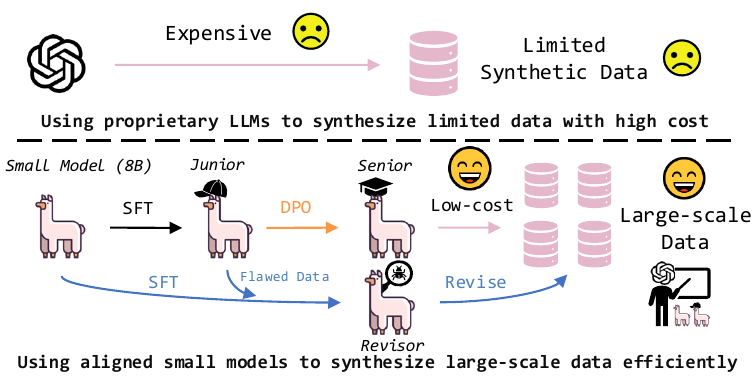}
	\caption{An illustration comparing the existing pipeline with our data synthesis framework.}
	\vspace{-2ex}
	\label{fig:introduction}
\end{figure}

However, most of these works solely use proprietary LLM like GPT-4 for data synthesis~\cite{E5mistral,Gecko}.
For example, E5$_\text{mistral}$ generates triplets of (query, positive document, hard negative document) for various embedding tasks from scratch.
While synthesizing embedding data without relying on existing corpora can yield more diverse examples, using black-box models can be extremely costly, especially given that this data often includes long documents.
A straightforward approach to reduce costs is to utilize small open-source models, which have proven effective for tasks such as mathematical reasoning~\cite{jiuzhang,bansal2024smaller,DBLP:journals/corr/abs-2407-18743}.
However, synthesizing embedding data often requires the generation of hard negatives -- documents that are similar to positive ones and are essential for learning nuanced embedding representations. 
These hard negatives are challenging for small models to synthesize, as they are difficult for language models to distinguish.
An early work explores the ability of small models for synthesizing embedding data~\cite{InParsv2}, but it uses small models to generate data directly without special tailoring for data synthesis, resulting in poor performance.

In this work, we propose to align open-source small models (8B) to synthesize large-scale high-quality embedding data.
Compared to existing methods that rely solely on expensive GPT-4, our approach can generate more data at a much lower cost. 
Our primary goal is to study the alignment of small models for synthesizing embedding data, which has been neglected by existing works.
Specifically, we aim to address the following research questions in this paper:

\noindent\textbf{RQ1}: How to align small models for synthesizing high-quality embedding data at scale?

\noindent\textbf{RQ2}: How do factors within the alignment framework affect the quality of synthetic data?

\noindent\textbf{RQ3}: Synthetic data is theoretically infinite. What is the scaling law for synthetic embedding data?

To shed light on \textbf{RQ1}, we design an alignment framework that trains small LLMs to efficiently \textbf{S}ynthesize large-scale su\textbf{PE}rior \textbf{E}mbedding \textbf{D}ata (\ours{}).
As illustrated in Figure~\ref{fig:introduction}, our framework consists of three key models: a junior generator for initial data synthesis, a senior generator for advanced data generation, and a data revisor for self-improvement. 
The goal is to distill knowledge from GPT-4 into these smaller models.
We first use GPT-4 to brainstorm task descriptions. 
However, since GPT-4 often generates hallucinations and data of specific domains (\eg, climate change)~\cite{chang2023examining}, we sample topics from the Open Directory Project to ensure diverse and balanced tasks.\footnote{\url{http://odp.org}: Open-source collection of web topics.} 
Based on these tasks, GPT-4 produces a small set of seed data, which we use to finetune the junior generator via supervised fine-tuning (SFT).
The junior generator produces root data, which is further evaluated by GPT-4 to produce signals that guide the preference optimization process, resulting in a senior generator.
The root data is also revised by GPT-4 to produce revision signals for training a data revisor.
Inspired by the idea of scaling inference compute for LLMs~\cite{llmmonkeys}, the revisor refines the synthetic data with minimal additional inference cost, enabling self-improvement.

As for \textbf{RQ2}, with these low-cost yet powerful data synthesis models ready, we are able to conduct extensive experiments to study the factors affecting the alignment.
We find that settings such as the base model used for alignment, the diversity of tasks, and the number of training samples can influence the quality of synthetic data.
For \textbf{RQ3}, we generate large-scale data using the efficient generators to reveal the scaling law.
We observe a log-linear relationship between the performance of the embedding model and the size of synthetic embedding data.

In summary, our contributions are as follows:
\begin{itemize}
    \item We design a framework to fine-tune small LMs (8B) for synthesizing large-scale data, achieving superior embedding performance with less than 1/10 of the GPT API calls required by E5$_\text{mistral}$.
    \item We comprehensively study how the factors within the alignment framework influence the quality of synthetic data.
    \item We investigate the scaling law of synthetic embedding data and reveal that the embedding model's performance follows a log-linear relationship with the data size.
\end{itemize}

\section{Related Work}

\noindent \textbf{Text Embedding}
Text embedding models have gained much attention in the era of deep learning.
Some existing models, such as SBERT~\cite{sbert}, E5~\cite{E5}, and BGE~\cite{bge}, attempt to produce general text embeddings for various tasks.
However, most of them require lots of labeled data.
In this work, we attempt to train a model with synthetic data.

\noindent \textbf{Large Language Models}
Though proprietary LLMs~\cite{gpt4,claude} are very powerful, invoking their APIs can be quite expensive and unaffordable for common usage. 
Many open-source LLMs have been released for more efficient language modeling, such as LLaMA~\cite{llama31} and Mistral~\cite{mistral}.
Some works attempt to improve the ability of LLMs for text embedding tasks, such as ad-hoc retrieval~\cite{repllama}, conversational retrieval~\cite{convaug}, and multilingual text embedding~\cite{E5mistral}.
Our work aims to use synthetic data to improve the LLM's ability of text embedding.

\begin{figure*}[!t]
	\centering
	\includegraphics[width=1.0\textwidth]{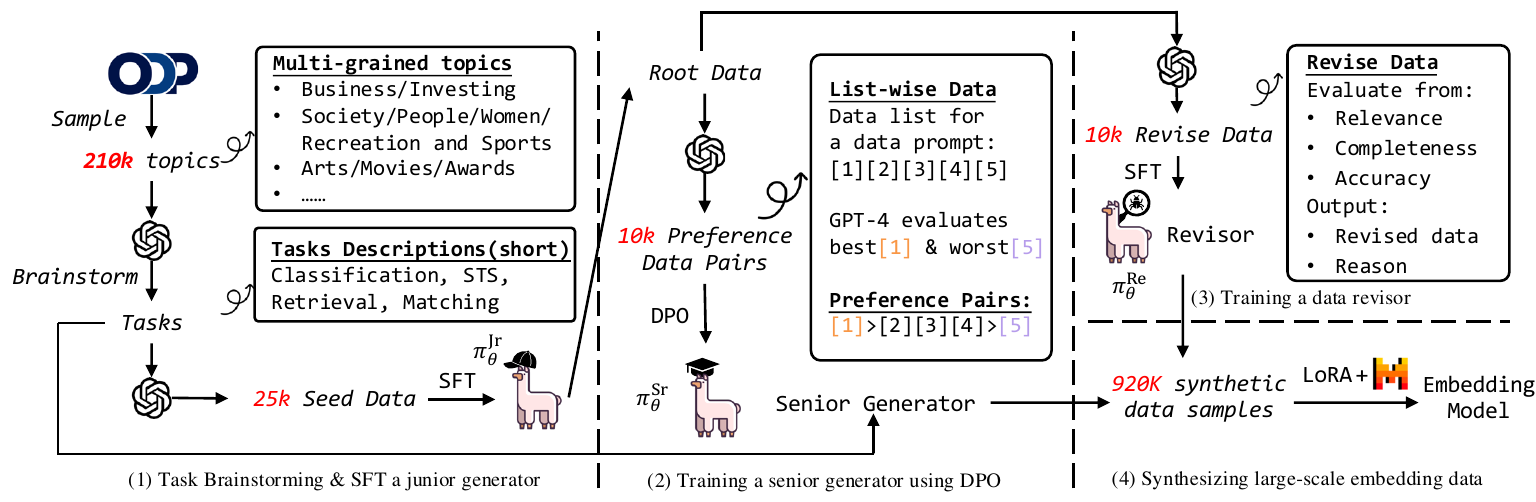}
	\caption{An overview of \ours{}. We align small LLMs (8B) to synthesize large-scale high-quality embedding data.}
	\vspace{-2ex}
	\label{fig:framework}
\end{figure*}

\noindent \textbf{Synthetic Data}
The generation of synthetic data have been studied by many researchers for various embedding tasks.
In early times, they have been used to produce pseudo labels and query/document expansions~\cite{doc2query,query2doc,promptgator}.
Using the ability of LLMs, synthetic data have been used for code generation~\cite{Textbooks,Qwen25Coder}, mathematical reasoning~\cite{DBLP:journals/corr/abs-2406-20094,glan,zhou2024enhancing,jiuzhang}, and text embedding~\cite{InParsv2,Prompt2Model,E5mistral,DBLP:journals/corr/abs-2407-12813,DBLP:conf/coling/PatwaF0CRM24,Gecko,jinav3}.
Though they have already shown great performance, most of these works heavily rely on black-box LLMs (\eg, E5$_\text{mistral}$~\cite{E5mistral} and Gecko~\cite{Gecko}) for data synthesis.
Our work aims to align small models for generating large scale text embedding data efficiently.

\section{Methodology: \ours{}}

In this section, we aim to answer \textbf{RQ1} using our alignment framework, \ours{}. As shown in Figure~\ref{fig:framework}, \ours{} consists of four stages: 
(1) GPT-4 is first used to generate diverse task descriptions based on multi-grained topics sampled from the ODP. A junior generator then distills knowledge from GPT-4 by training on a small set of seed data. 
(2) The junior generator synthesizes root data, which GPT-4 uses to produce preference signals. These signals are used to train a senior generator through preference optimization. 
(3) The root data is also evaluated by GPT-4 to produce revised data for finetuning a data revisor. 
(4) Finally, the senior generator synthesizes large-scale embedding data, and the revisor refines them into high-quality data for training the embedding model.

\subsection{Preliminaries}

 Many works have tried to generate synthetic data using modern LLMs for downstream tasks finetuning.
Following E5$_\text{mistral}$~\cite{E5mistral}, in order to synthesize data for training an embedding model, we generate data for four kinds of tasks: classification (long-short match), semantic textual similarity (STS), retrieval (short-long match), and text matching (short-short and long-long match).
For simplicity, we will denote the data synthesis prompts as a set $P$ without distinction.\footnote{Since our research focus is how to align small models to synthesize embedding data efficiently rather than adjusting prompts for the synthesis process, we will follow the task types and prompt templates in E5$_\text{mistral}$.}
We use GPT-4 to brainstorm a pool of candidate tasks $T$ as instructions.
With a prompt $p \in P$ and a task instruction $t \in T$, an LLM $\pi_{\theta}$ can synthesize an embedding data sample $d \sim \pi_{\theta} (d \mid p, t) $.
Each data example is a triplet of (query, positive document, hard negative document).
For example, for a classification task, the query is a long text and documents are short labels.
More information on the structure of these data can be found in Appendix~\ref{appendix: example}.

\subsection{Aligning Small Models for Synthesizing Embedding Data}

Most existing approaches that synthesize embedding data suffer from the high cost of heavily relying on proprietary LLMs.
We aim to align small models that can generate large-scale embedding data effectively and efficiently.

\subsubsection{Task Brainstorming} 
Synthesizing embedding data from scratch can be quite challenging since these data are often long and complex.
We first generate a pool of candidate tasks as instructions for LLMs to further generate concrete data.
Since these task descriptions are very short (about 10 words) and need to be high-quality, we use GPT-4 to brainstorm them.
Furthermore, we sample multi-grained topics from open directory project (ODP) and specify one topic for each brainstorming prompt to mitigate the hallucination and extract more diverse knowledge from GPT-4~\cite{chang2023examining}.
For example, we prompt GPT-4 as "\textit{Brainstorm a list of potentially useful text retrieval tasks for the topic: \{topic\}.}".\footnote{Due to space limitation, we will not present full prompts in this section. The complete prompts are in Appendix~\ref{appendix: prompt}.}
Then we will get a diverse set of task descriptions and generate embedding data conditioned on them.

\subsubsection{Training a Junior Generator}
Proprietary LLMs such as GPT-4 have been proven to generate high-quality embedding data~\cite{E5mistral,Gecko}.
However, it can be expensive if we generate large-scale embedding data solely using GPT-4.
Our goal is to distill the data synthesis capability of GPT-4 into small models that can synthesize large-scale data at low cost.

We first use GPT-4 to generate a small set of seed data $D_{\text{seed}}\sim \pi^{\text{GPT-4}}_{\theta}  (D_{\text{seed}} \mid P, T) $.
The constructed training data for SFT is $D_{\text{SFT}}=\{p_i, t_i, d_i\}_{i=1}^N$.
To distill knowledge from GPT-4, we apply a standard Supervised Fine-tuning (SFT) objective to initialize our junior generator $\pi^{\text{Jr}}_{\theta}$: 
\begin{equation}
\mathcal{L}(\theta^{\text{Jr}}) = -\sum\nolimits_{(p_i, t_i, d_i)\in\mathcal{D}_{\text{SFT}}} \log \mathbb{P}_{\theta}(d_i \mid p_i, t_i),
\end{equation}
where $\theta^{\text{Jr}}$ denotes the parameters of our junior generator. 
We aim to train a small model with basic capability of synthesizing embedding data given various prompt templates and task instructions.

\subsubsection{Further Training Using Preference Optimization}

Although our junior generator can already generate embedding data of decent quality, we still want to boost its ability.
Preference optimization~\cite{ppo} is a popular way to be performed on a model for further training after SFT~\cite{selfplay,DBLP:journals/corr/abs-2405-21040}.
Since our goal is to perform optimization on $\pi^{\text{Jr}}_{\theta}$, we use GPT-4 to produce preference signals based on the data generated by $\pi^{\text{Jr}}_{\theta}$ itself.

Specifically, $\pi^{\text{Jr}}_{\theta}$ generates a list of embedding data given each prompt, formatting a set of root data $D_{\text{root}}\sim \pi^{\text{Jr}}_{\theta}  (D_{\text{root}} \mid P, T) $.
As illustrated in Figure~\ref{fig:framework}, GPT-4 evaluates the best and the worst data in each data list and constructs preference pairs accordingly.
We prompt GPT-4 as: "\textit{Your mission is to judge which data this language model generates fits the prompt most and which fits worst, and explain your judgment.}". 
In this work, we perform Direct Preference Optimization (DPO)~\cite{dpo} because it is a popular and low-cost method. 
The formatted training set for DPO is $D_{\text{DPO}}=\{p, t, d_w, d_l, \}$, where $d_w$ and $d_l$ are the winning and losing one, respectively.
Then, we apply the standard DPO on our junior generator:
\begin{equation}
\begin{split}
& \mathcal{L}_{\text{DPO}}(\pi^{\text{Jr}}_{\theta};\pi_{\text{ref}}) = \\ 
& -\mathbb{E}_{(p, t, d_w, d_l)\sim \mathcal{D}} \left[\log \sigma \left(\beta \log \frac{\pi^{\text{Jr}}_{\theta}(d_w\mid x)}{\pi_{\text{ref}}(d_w\mid x)} \right. \right. \\
& \left. \left. - \beta \log \frac{\pi^{\text{Jr}}_{\theta}(d_l\mid x)}{\pi_{\text{ref}}(d_l\mid x)}\right)\right], 
\end{split}
\end{equation} 
where $\pi_{\text{ref}}$ is the reference model set as $\pi^{\text{Jr}}_{\theta}$ in the beginning and remains frozen, $\sigma$ is the sigmoid function, and $\beta$ controls how much DPO focus on $\pi_{\text{ref}}$.
After this, we manage to obtain a senior generator $\pi^{\text{Sr}}_{\theta}$ that can synthesize higher-quality data since it has learned about how to make better choices given a data synthesis prompt.

\subsubsection{Training a Data Revisor}

Scaling the inference compute of LLMs has been a popular way to boost the LLM's performance from the inference side~\cite{llmmonkeys}.
Inspired by this, we employ another small model to refine our synthetic data.
This allows us to further improve data quality with only a small increase in inference cost, as the revisor model is also small.
Specifically, we train an additional LLM to serve as the data revisor, identifying and refining potential flaws in the synthetic data.

Specifically, to boost the efficiency of the alignment process, we reuse $D_{\text{root}}$ to produce revised data.
This allows us to train both $\pi^{\text{Sr}}_{\theta}$ and the revisor $\pi^{\text{Re}}_{\theta}$ simultaneously.
GPT-4 produces data revision signals by evaluating the root data from three key aspects: (1) its relevance to the task, (2) its completeness based on the requirements in the prompt, (3) the accuracy of its factual content.
The revised data is $D^{\text{re}}_{\text{root}}\sim \pi^{\text{GPT-4}}_{\theta}  (D^{\text{re}}_{\text{root}} \mid P, T, D_{\text{root}}) $ and the data for SFT is $D^{\text{re}}_{\text{SFT}}=\{p_j, t_j, d^{\text{root}}_j, d^{\text{re}}_j\}_{j=1}^M$.
Similarly, a standard SFT approach is performed on an unaligned small LM:
\begin{align}
\mathcal{L}(\theta^{\text{Re}}) &= -\sum\nolimits_{(x_j, d^{\text{re}}_j)\in\mathcal{D}^{\text{re}}_{\text{SFT}}} \log \mathbb{P}_{\theta}(d^{\text{re}}_j \mid x_j), \notag \\
x_j &= (p_j, t_j, d^{\text{root}}_j),
\end{align}
where $\theta^{\text{Re}}$ denotes the parameters of our revisor.

\begin{table*}[!t]
\small
\setlength{\tabcolsep}{3.5pt}
\renewcommand{\arraystretch}{1.2}
\begin{tabular}{lcc|cccccccc}
\toprule
                         & Synthesis Model          & \# FT. Data & Class. & Clust. & Pair. & Rerank. & Retr. & STS  & Summ. & Avg. \\ \hline
\multicolumn{11}{l}{\textit{Zero-shot Models (w/ synthetic data only)}}                                                                  \\ \hline
$\text{Mistral}_\text{llama3}$       & llama3-8B-instruct & 230K         & 76.8   & 48.0   & 79.8   & \underline{59.5}  &  44.2   & 79.7  & 31.5 & 61.0 \\
$\text{Mistral}_\text{gpt-4o}$        & gpt-4o              & 230K         & 77.7   & 47.7   & 83.9  & 58.7    & 46.7  & 80.9 & 30.7  & 62.2 \\
Gecko$_\text{1b-768}$        & black-box            & 6.6M         & 70.3   & 46.8   & \underline{86.2}  & 57.6    & \textbf{53.2}  & \textbf{83.1} & \textbf{32.2}  & 62.6 \\
E5$_\text{mistral-7b}$            & gpt-3.5(25\%)+gpt-4(75\%)              & 500K$^\text{m}$        & \underline{78.2}   & \textbf{50.5}   & {86.0}  & 59.0    & 46.9  & 81.2 & \underline{31.9}  & \underline{63.1} \\
\ours{} (Ours) & llama3-8B-aligned       & 920K  & \textbf{78.3}        &  \underline{48.6}      &  \textbf{86.3}      &  \textbf{59.8}     &  \underline{48.1}       &  \underline{82.6}     &  31.7    &   \textbf{63.4}        \\ \hline

\multicolumn{11}{l}{\textit{Supervised Models (w/ synthetic data + labeled data)}}                                                                            \\ \hline
GTR$_\text{xxl}$              & -                   & 662K            & 67.4   & 42.4   & 86.1  & 56.7    & 48.5  & 78.4 & 30.6  & 59.0 \\
$\text{GTE}_\text{large}$            & -                   & 3M            & 73.3   & 46.8   & 85.0  & 59.1    & 52.2  & 83.4 & \underline{31.7}  & 63.1 \\
text-embedding-3$_\text{large}$   & -                   & -            & 75.5   & 49.0   & 85.7  & 59.2    & 55.4  & 81.7 & 29.9  & 64.6 \\
jina-embeddings-v3       & -                   & -            & \textbf{82.6}   & 45.3   & 84.0  & 58.1    & 53.9  & \textbf{85.8} & 29.7  & 65.5 \\
Gecko$_\text{1b-768}$         & black-box            & >6.6M         & \underline{81.2}   & 47.5   & 87.6  & 58.9    & 55.7  & 85.1 & \textbf{32.6}  & 66.3 \\
E5$_\text{mistral-7b}$               & gpt-3.5(25\%)+gpt-4(75\%)                & 1.8M         & 78.5   & \textbf{50.3}   & \textbf{88.3}  & \underline{60.2}    & \textbf{56.9}  & 84.6 & 31.4  & \textbf{66.6} \\
\ours{} (Ours) & llama3-8B-aligned          & 2.2M         & 78.4       &  \underline{49.3}      &  \underline{88.2}     &  \textbf{60.8}       &  \underline{56.5}     &  \underline{85.5}    &  31.1     & \underline{66.5} \\
\bottomrule
\end{tabular}
\caption{Results on MTEB benchmark, including 56 tasks of 7 types: Classification (Class.), Clustering (Clust.), Pair Classification (Pair.), Reranking (Rerank.), Retrieval (Retr.), Semantic Textual Similarity (STS), and Summarization (Summ.). ``Synthesis Model'' denotes the LLM used for generating synthetic data. ``\# FT. Data'' denotes the data amount used for finetuning the embedding models. ``500K$^\text{m}$'': E5$_\text{mistral-7b}$ is a multilingual model, it synthesized 190K English samples plus 310K samples of other languages. The best performances are in bold and the second-best performances are underlined.}
\label{tab:main_results}
\end{table*}

\subsection{Finetuning Embedding Model Using Synthetic Data}

With our aligned senior generator $\pi^{\text{Sr}}_{\theta}$ and revisor $\pi^{\text{Re}}_{\theta}$ ready, we are able to generate high-quality synthetic embedding data at scale.
Specifically, $\pi^{\text{Sr}}_{\theta}$ first generates a large set of synthetic data $D_{\text{syn}}\sim \pi^{\text{Sr}}_{\theta}  (D_{\text{syn}} \mid P, T) $.
Then $\pi^{\text{Re}}_{\theta}$ revises them into high-quality data $D^{\text{re}}_{\text{syn}}\sim \pi^{\text{Re}}_{\theta}  (D^{\text{re}}_{\text{syn}} \mid P, T, D_{\text{syn}}) $.
For efficiency, we avoid iterative improvements and perform the revision in a single pass.

Following the common approach of task-specific fine-tuning~\cite{bge,E5mistral}, an instruction template is applied on each query within $D^{\text{re}}_{\text{syn}}$ as: $ q^i = \text{Instruct:} \{t\}~\textbackslash n~\text{Query:} \{q\}$, where $q^i$ is the original query $q$ with task description.
We do not apply this template on the document side for pre-building the index.
We append an [EOS] token to each $q^i$ and document $d$. Each output of the last layer [EOS] is taken as the representation $\mathbf{q}^i$ and $\mathbf{d}$.
To train the embedding model, we apply a standard contrastive learning objective:
\begin{eqnarray}
\mathcal{L}_{\text{CL}} = -\log\frac{\phi(\mathbf{q}^i,\mathbf{d}^+)}{\phi(\mathbf{q}^i,\mathbf{d}^+) + \sum_{{d}^-\in\mathcal{N}}{\phi(\mathbf{q}^i,\mathbf{d}^-)}},
\end{eqnarray}
where $\mathcal{N}$ represents negative documents, $\phi(\cdot) = \exp(\text{cos}(\cdot) / \tau)$, ${\rm cos}(\cdot)$ denotes cosine similarity, and $\tau$ is a temperature hyperparameter.

\section{Experiments}

\subsection{Experimental Setup}
\label{subsec:setup}

\ours{} synthesizes 920K embedding data samples in total for training after MinHash deduplication.
The proprietary LLM used for knowledge distillation is \textit{GPT-4o-2024-05-13}.
The base model we use to train our generators is LLaMA-3-8B~\cite{llama3}.
We test our finetuned embedding model on the MTEB benchmark~\cite{mteb}.
This benchmark contains 7 kinds of 56 English embedding tasks: classification (12), clustering (11), pair classification (3), reranking (4), retrieval (15), semantic textual similarity (10) and summarization (1).
The synthetic data proportion of our four embedding task types, \ie, classification, STS, retrieval, and text matching is 7:7:7:2.
For fair comparisons to E5$_\text{mistral}$, we train Mistral-7B-v0.1~\cite{mistral} as our embedding model and use the same labeled data for ``\textit{Supervised Models}'' setting.
We use LoRA~\cite{lora} to finetune our embedding model.

In addition to existing baselines that consists of OpenAI's text-embedding-3\footnote{\url{https://platform.openai.com/docs/guides/embeddings}}, GTR~\cite{gtr}, GTE~\cite{gte}, jina-embeddings-v3~\cite{jinav3}, Gecko~\cite{Gecko}, and E5$_\text{mistral-7b}$~\cite{E5mistral}, we also implement two baselines finetuned on synthetic data only.
In particular, we use llama3-8B-instruct and gpt-4o to synthesize 230K embedding data using the same synthesis prompts and data proportion of \ours{}.
Then we finetune Mistral-7B-v0.1 with these data to produce two baselines: $\text{Mistral}_\text{llama3}$ and $\text{Mistral}_\text{gpt-4o}$.

More details about the synthetic data, implementation details, and prompts can be found in Appendix~\ref{appendix: syndata}, \ref{appendix: implementation}, and~\ref{appendix: prompt}, respectively.

\subsection{Main Results}

The results are presented in Table~\ref{tab:main_results}.
\ours{} achieves the best performance in the zero-shot setting and the second-best performance in the supervised setting.
This demonstrates the effectiveness of our framework, as \textbf{\ours{} can generate large-scale high-quality data using the smallest language model}. 
These results address \textbf{RQ1}, confirming that \ours{} is an effective way to align small models for synthesizing large-scale embedding data.
Furthermore, we can make these observations:
(1) Comparing to $\text{Mistral}_\text{llama3}$, \ours{} improves its performance greatly. 
This demonstrates that our alignment framework enables a base small model to synthesize higher-quality data than its instruct-tuned version. Additionally, as shown in Table~\ref{tab:ablation}, \ours{} with just 230K data examples also outperforms Mistral$_\text{llama3}$.
(2) Intriguingly, \ours{} outperforms E5$_\text{mistral-7b}$ in the zero-shot setting but slightly underperforms in the full-data setting. 
We attribute this to the fact that, while our synthetic data is more diverse and covers a broader range of scenarios, E5$_\text{mistral-7b}$'s data is structurally closer to labeled data, as it is generated by the powerful but costly GPT.
(3) Gecko performs well on some certain types of embedding tasks. 
We believe this is because Gecko uses a black-box model to generate a large set of synthetic data (6.6M), potentially covering more task types than both \ours{} and E5$_\text{mistral-7b}$.

\subsection{RQ2. Alignment Analysis}

In this section, we will look deeper into \ours{} and provide comprehensive analysis of how each factor influences the synthetic data. 
For efficient analysis, we synthesize 230K embedding data using the same data proportion of \ours{} for each model and perform zero-shot evaluation on MTEB.

\subsubsection{Ablation Study}

\begin{table}[t!]
    \centering
    \small
    \begin{tabular}{p{0.25\textwidth}cc}
    \toprule
         Model & Avg. on MTEB  \\
        \midrule
        \ours{} (230K synthetic data) & \textbf{63.2}  \\
        \quad w/ only SFT (only $\pi^{\text{Jr}}_{\theta}$)& 62.6    \\
        \quad w/o. DPO ($\pi^{\text{Jr}}_{\theta}$+$\pi^{\text{Re}}_{\theta}$)& 62.8   \\
        \quad w/o. Data Revisor (only $\pi^{\text{Sr}}_{\theta}$) & 62.9    \\
    \bottomrule
    \end{tabular}
    \caption{Performances of ablated models on MTEB.}
    \label{tab:ablation}
\end{table} 

To evaluate each component of \ours{}, we first conduct ablation experiments on our alignment framework.
The results are presented in Table~\ref{tab:ablation}.
We can make the following observations:
(1) $\pi^{\text{Jr}}_{\theta}$ itself can already synthesize embedding data of decent quality (62.6), which demonstrates the effectiveness of our aligned junior generator.
(2) ``\ours{} w/o. DPO'', \ie, only $\pi^{\text{Jr}}_{\theta}$ and $\pi^{\text{Re}}_{\theta}$ causes performance decreasing.
This demonstrates our DPO training process can further enhance the synthesis ability of $\pi^{\text{Jr}}_{\theta}$.
(3) The performance drops after discarding $\pi^{\text{Re}}_{\theta}$. 
This shows revising the synthetic data with our data revisor can enhance the data quality by introducing a little more inference compute.

\subsubsection{Task Brainstorming}

\begin{table}[t!]
    \centering
\small
\renewcommand{\arraystretch}{1.2}
\begin{tabular}{p{0.27\textwidth}cc}
\toprule
Topic \& Task                   & Avg. on MTEB \\
\midrule
$\pi^{\text{Jr}}_{\theta}$ (1 task per topic \& truncation) & \textbf{62.6}        \\
\hline
\multicolumn{2}{l}{\textit{\# Tasks per topic}}         \\
\hline
3 tasks per topic              & 61.6        \\
5 tasks per topic              & 60.9        \\
\hline
\multicolumn{2}{l}{\textit{Topic granularity}}          \\
\hline
Specific topic (w/o. truncation)                  & 61.8  \\
\bottomrule
\end{tabular}
\caption{Performances of models with different settings of task brainstorming on MTEB. For efficient test, the models have only been through SFT with 230K data.}
\label{tab:topic}
\end{table}

To mitigate hallucination and introduce diversity to LLMs, we propose to use GPT-4 to brainstorm a candidate pool of task descriptions with multi-grained topics before we synthesize specific data.
To study the influence of topic diversity and coverage, we perform experiments from two aspects and present the results in Table~\ref{tab:topic}: 
(1) The number of tasks per topic.
For each topic sampled from ODP, we generate 1, 3, and 5 tasks.
We find that the performance of $\pi^{\text{Jr}}_{\theta}$ drops greatly when we generate more tasks per topic. This demonstrates that the diversity of tasks is important for the quality of synthetic data.
(2) The granularity of topics.
The sampled topics are multi-grained and we truncate those extremely specific topics to a maximum depth of 4.
Without truncation, those topics will produce tasks harming the generalization of \ours{}.

\subsubsection{Junior Generator $\pi^{\text{Jr}}_{\theta}$}
In this section, we will look into our SFT process and discuss the factors that may influence $\pi^{\text{Jr}}_{\theta}$:

\begin{figure*}[tbp]
\centering
\includegraphics[width=0.95\textwidth]{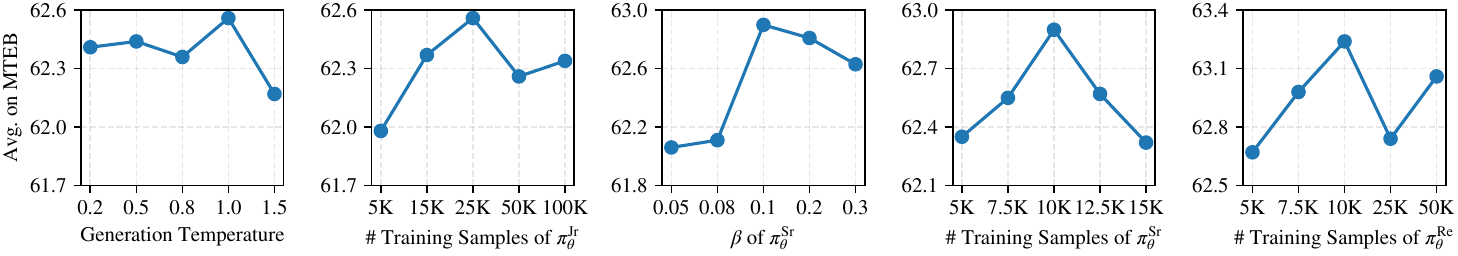}
\vspace{-2pt}
\caption{Performances of \ours{} (230K data for efficient test) with different settings of the alignment pipeline.}\label{fig:factors} 
\end{figure*}

\begin{table}[t!]
    \centering
    \small
    \begin{tabular}{p{0.3\textwidth}cc}
    \toprule
        Base Model for $\pi^{\text{Jr}}_{\theta}$ & Avg. on MTEB  \\
        \midrule
        LLaMA-3-8B~\cite{llama3} (Ours)  & \textbf{62.6}  \\
        LLaMA-2-7B~\cite{llama2} &  62.4  \\
        Gemma-7B~\cite{gemma} & 62.3   \\
        Qwen-2.5-7B~\cite{qwen25} & 62.5   \\
    \bottomrule
    \end{tabular}
    \caption{Performances of $\pi^{\text{Jr}}_{\theta}$ with different base models.}
    \label{tab:base_model}
\end{table} 

\noindent \textbf{Base LLM.}
The base model that we train into our synthesis LLM is directly related to the data quality.
To study this, we apply our SFT pipeline on several other base LLMs.
From the results in Table~\ref{tab:base_model}, we can observe that all LLMs can synthesize embedding data of decent quality with our SFT pipeline.
This shows the effectiveness and applicability of our designed alignment process again.
Besides, $\pi^{\text{Jr}}_{\theta}$ trained on LLaMA-3-8B achieves the best performance, which is consistent with its superior language modeling ability.
This means we can easily boost the quality of synthetic data by applying \ours{} on more advanced open-source LLMs.

\noindent \textbf{The generation temperature.}
Temperature is a crucial hyperparameter that controls the randomness of the text generation process.
We set the generation temperature of $\pi^{\text{Jr}}_{\theta}$ in the range of [0.2, 1.5], and present the performances on MTEB in the left part of Figure~\ref{fig:factors}. 
Due to space limitations, we only show results for five values (this policy will be followed in the subsequent displays).
We can observe that the performance of $\pi^{\text{Jr}}_{\theta}$ first increases then drops.
This phenomenon indicates a trade-off:
If the temperature is too low, the synthetic data will lack diversity.
However, the LLM may generate data that do not follow the required structure and guidelines if the temperature is too high.

\noindent \textbf{The number of training samples.}
In our training process of $\pi^{\text{Jr}}_{\theta}$, we use GPT-4 to produce signals for knowledge distillation.
This raises a question: how many samples should we use for finetuning the generator? Is it the more the better?
We study this question by set the number of training samples of $\pi^{\text{Jr}}_{\theta}$ in the range of [5K, 100K].
As shown in the middle left part of Figure~\ref{fig:factors}, a small set of training samples can already train a decent generator using our SFT pipeline, which validates its effectiveness again.
However, too many training samples will harm the language modeling ability of the LLM.

\subsubsection{Senior Generator $\pi^{\text{Sr}}_{\theta}$}

We propose to further train the junior generator with DPO into a more powerful synthesis model $\pi^{\text{Sr}}_{\theta}$.
In this part, we will look into this process from these aspects:

\noindent \textbf{The hyperparameter $\beta$.}
When performing DPO on $\pi^{\text{Jr}}_{\theta}$, we aim to improve its performance by directly optimizing for preference signals produced by GPT-4.
$\beta$ is the hyperparameter used to control the trade-off between aligning the model to preference signals and avoiding over-optimization that may degrade performance on the original task.
To study it empirically, we set $\beta$ in the range of [0.05, 0.3].
As presented in the middle part of Figure~\ref{fig:factors}, \ours{}'s performance increases to an optimal value when $\beta=0.1$ then drops.
This validates the trade-off:
A high $\beta$ controls $\pi^{\text{Sr}}_{\theta}$ to stay close to the reference model ($\pi^{\text{Jr}}_{\theta}$), ensuring it doesn't drift too much, while a low $\beta$ encourages stronger adaptation to the preference signals, but at the risk of overfitting.

\noindent \textbf{The number of training samples.}
Similar to the SFT process, we can raise a question: how many preference data pairs we should use to align $\pi^{\text{Sr}}_{\theta}$?
We study this question by setting the number of training samples for $\pi^{\text{Sr}}_{\theta}$ in the range of [5K, 15K].
From the results in the middle right part of Figure~\ref{fig:factors}, we can observe that finetuning $\pi^{\text{Jr}}_{\theta}$ using DPO needs fewer data that the SFT process.
This is consistent with previous studies that pairwise signals of outputs (preferences) are more informative per instance than standard supervised data.
We also notice that the performance drops when we use too many preference signals.
This indicates that overfitting the junior generator will harm its ability of following basic guidelines and instructions.

\subsubsection{Data Revisor $\pi^{\text{Re}}_{\theta}$}

\noindent \textbf{The number of training samples.}
\ours{} further enhances the quality of synthetic embedding data using a data revisor.
GPT-4 evaluates the root data synthesized by $\pi^{\text{Jr}}_{\theta}$ from multi-grained aspects and produces data revision signals to finetune $\pi^{\text{Re}}_{\theta}$.
$\pi^{\text{Re}}_{\theta}$ revises the synthetic data generated by $\pi^{\text{Sr}}_{\theta}$ to take a reflection at them and boost their quality.
To study the influence of the number of the revision signals used for aligning the revisor, we set it in the range of [5K, 50K].
As shown in the right part of Figure~\ref{fig:factors}, we can observe a similar pattern as the training of $\pi^{\text{Jr}}_{\theta}$.
This is consistent with their training protocol that they are both aligned by SFT.
However, it takes fewer training data to finetune $\pi^{\text{Re}}_{\theta}$ than $\pi^{\text{Jr}}_{\theta}$.
This is because that it is easier to revise a data sample of decent quality than synthesize one from scratch.

\subsection{RQ3. Scaling Synthetic Embedding Data}

\begin{figure}[t]
\centering
\includegraphics[width=0.45\textwidth]{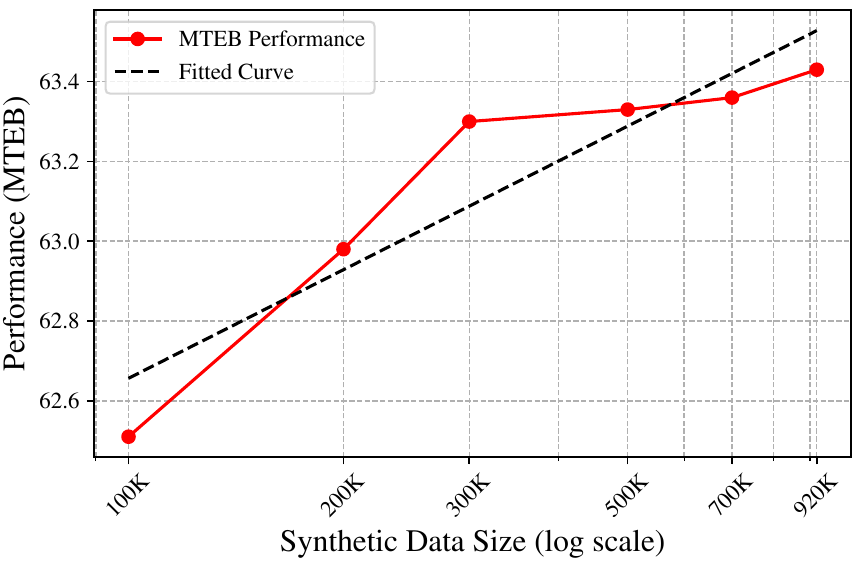}
\vspace{-2pt}
\caption{Scaling laws for model performance in relation to synthetic embedding data size on MTEB.}
\label{fig:scaling_law} 
\end{figure}

In the era of LLMs, models are often trained on billions or even trillions of data points. 
This raises a key question: does increasing training data always lead to better performance? 
Some existing works has explored this through scaling laws in areas like language modeling~\cite{LMscale} and dense retrieval~\cite{DRscale}.
However, these works primarily focus on scaling the labeled data or existing corpora.

Synthetic data, which are theoretically unlimited, remains an underexplored area for scaling laws~\cite{synsurvey}. This is a non-trivial problem because:
(1) The distribution of synthetic data differs from that of labeled data~\cite{llmsyntale}.
(2) Generating large-scale synthetic data with black-box LLMs to study scaling laws can be costly.
With the efficient data synthesis capabilities of \ours{}, we are able to generate large-scale embedding data and analyze the corresponding scaling law. As shown in Figure~\ref{fig:scaling_law}, we observe a log-linear relationship between the embedding model’s performance and the size of the synthetic data.
This scaling law offers key insights for future works:
(1) The log-linear trend enables researchers to predict performance improvements from synthesizing more data.
(2) It guides trade-offs by showing diminishing returns—beyond a certain point, additional data yields marginal improvement, making further investment in data synthesis less valuable.

\subsection{Cost Analysis}

\begin{table}[t!]
\centering
\small
\renewcommand{\arraystretch}{1.2}
\begin{tabular}{lcc}
\toprule
Model      & GPT API Calls & GPT Token Usage \\ \hline
E5$_\text{mistral}$   & 500K                    & 180M            \\ 
\ours{}    & 45K              & 32M              \\ 
\bottomrule
\end{tabular}
\caption{Cost comparison between \ours{} and E5$_\text{mistral}$ in terms of GPT API calls and token usage.}
\label{tab:cost-analysis}
\end{table}

In this section, we analyze the cost of our alignment framework, \ours{}. 
The cost is reported from two aspects: GPT API calls (the number of invoking times) and GPT token usage. 
We omit the task brainstorming process, as the task descriptions are very short compared to the embedding data, and we also neglect the cost of deploying the aligned generators since they are very small.

Specifically, \ours{} costs 25K (SFT $\pi^{\text{Jr}}_{\theta}$) + 10K (DPO $\pi^{\text{Sr}}_{\theta}$) + 10K (SFT $\pi^{\text{Re}}_{\theta}$) = \textbf{45K} GPT API calls.
As for GPT token usage it costs 10M (SFT $\pi^{\text{Jr}}_{\theta}$) + 12M (DPO $\pi^{\text{Sr}}_{\theta}$) + 10M (SFT $\pi^{\text{Re}}_{\theta}$) = \textbf{32M}.

For a more staightforward understanding, we compare these costs with the synthesis process of E5$_\text{mistral}$, which solely uses GPT to synthesize data. 
It requires 500K API calls and consumes 180M GPT tokens~\cite{E5mistral}. 
The comparison, shown in Table~\ref{tab:cost-analysis}, highlights that \ours{} is significantly more efficient, requiring only less than 1/10 of the GPT-4 API calls and about 1/6 of the tokens to align small open-source models for synthesizing large-scale data efficiently and effectively.
\section{Conclusion}

In this work, we propose a framework \ours{} that aligns small models for the efficient and effective synthesis of embedding data. 
Through supervised finetuning, preference optimization, and self-improvement, small models can also synthesize high-quality embedding data at scale.
Additionally, we comprehensively investigate how various factors within the alignment pipeline influence data quality.
We reveal the scaling law of synthetic embedding data, demonstrating a log-linear relationship between the performance of the embedding model and the size of the synthetic data.
\clearpage

\section*{Limitations}

Our work still have several limitations that we plan to address in future works:

\begin{enumerate}
    \item The training signals we produce may be improved in the future. Although GPT-4o is already a very powerful LLM, it still can not perfectly interpret the guidelines and requirements in our prompts. For example, some of the long hard negative documents are too close to the positive ones.
    \item Our senior generator is trained by DPO. More advanced preference optimization approaches such as step-DPO will be utilized.
    \item The base models used for data synthesis and embedding model can be improved. For fair comparisons to baselines, we train Mistral-7B-v0.1 as our embedding model. In future works, we plan to use more advanced LLMs to boost our model's performance.
    \item We do not fit a function for the scaling law we reveal for synthetic embedding data. In future work, we will explore a power-law function that can represent the scaling relationship we find in this paper.
    
\end{enumerate}


\appendix

\clearpage

\section*{Appendix}

\section{Details about Synthetic Data}
\label{appendix: syndata}

\begin{table}[h]
\centering
\begin{tabular}{lcc}
\toprule
Synthetic Task Type           &  \# Examples  \\ \midrule
Classification (long-short match)    & 245,947              \\
Semantic textual similarity (STS)            & 294,388           \\
Retrieval (short-long match)         & 303,424             \\ 
Text matching (short-short match)            & 39,954           \\
Text matching (long-long match)            & 36,702          \\
\bottomrule
\end{tabular}
\caption{Statistics of the synthetic data (after MinHash) used for finetuning the embedding model.}
\label{tab:syndata_statistics}
\end{table}

In this section, we will look into the detailed information and statistics of generated synthetic embedding data.
The statistics is presented in Table~\ref{tab:syndata_statistics}.
We first generates a raw synthetic dataset of 1.15M examples following the data proportion in Section~\ref{subsec:setup}.
And after MinHash deduplication, there are 920,415 data left in total.

\section{Implementation Details}
\label{appendix: implementation}

In this part, we delve into the details about the implementation of \ours{}.
Specifically, we finetune LLaMA-3-8B as data synthesis models and Mistral-7B-v0.1 as our embedding model.
For the SFT process of $\pi^{\text{Jr}}_{\theta}$, the learning rate is 1e-4 and the batch size is 16. 
As for the DPO process of $\pi^{\text{Sr}}_{\theta}$, the learning rate is 1e-5, beta $\beta$ is set as 0.1, and the batch size is 16. 
For the SFT process of $\pi^{\text{Re}}_{\theta}$, the learning rate is 5e-6 and the batch size is 24. 

For the data generation, we set the temperature as 1.0 for all data synthesis except 0.0 for producing the preference signal.
The top\_p is set as 1.0.

For the training of our embedding model, we use LoRA with rank 16 and DeepSpeed ZeRO-3.
We set the batch size as 1,536 using 16 40G A100 and fp16.
For the training data, we use a combination of synthetic data and a collection of 13 public datasets.
These labeled datasets used for finetuning are the same as those in E5$_\text{mistral}$.

For the instructions we used for the training and evaluation datasets (MTEB), please refer to the original paper of E5$_\text{mistral}$~\cite{E5mistral}.

\section{Prompts}
\label{appendix: prompt}

The prompts we used in our work can be categorized into two kinds: prompts used for generating synthetic data and aligning data generators.

\subsection{Data Generation}

Since our work focuses on the alignment of small models for synthesizing large-scale embedding data, we reuse most of the data generation prompts and data structures of E5$_\text{mistral}$~\cite{E5mistral}.
For task brainstorming, we adjust those prompts to fit the sampled topic by appending ``for the topic: \{topic\}'' after each ``Brainstorm a list of potentially useful xxx tasks''.
For the synthesis of STS data we change its prompt to fit the sampled topics as follows:

\begin{prompt}[title={Prompt: Synthesizing STS Data}, label=prompt:sts]

Write a \textcolor{blue}{\{sentence, phrase, passage\}} triple for the topic: \textcolor{blue}{\{topic\}} with varying semantic similarity scores in JSON format. The semantic similarity score ranges from 1 to 5, with 1 denotes least similar and 5 denotes most similar.\\ \\ Please adhere to the following guidelines:\\ - The keys in JSON are "S1", "S2", and "S3", the values are all strings in English, do not add any other keys.\\ - There should be some word overlaps between all three \textcolor{blue}{\{sentence, phrase, passage\}}s.\\ - The similarity score between S1 and S2 should be \textcolor{blue}{\{4, 4.5, 5\}}.\\ - The similarity score between S1 and S3 should be \textcolor{blue}{\{2.5, 3, 3.5\}}.\\ - The \{sentence, phrase, passage\}s require \textcolor{blue}{\{elementary school, high school, college\}} level education to understand and should be diverse in terms of topic and length.\\ \\ Your output must always be a JSON object only with three keys "S1", "S2" and "S3", do not explain yourself or output anything else. Be creative!

\end{prompt}

\subsection{Generator Alignment}

In this part, we will shed light on the prompts we use to generate the signals for knowledge distillation.
For the SFT of $\pi^{\text{Jr}}_{\theta}$, the training data are sampled from the synthesis of $\text{Mistral}_\text{gpt-4o}$.
For the DPO of $\pi^{\text{Sr}}_{\theta}$, we prompt GPT-4 to produce preference data as:

\begin{prompt}[title={Prompt: Generating Preference Data}, label=prompt:dpo]

A language model has been given a prompt: \textcolor{blue}{\{data prompt\}}

The output list of it is: \textcolor{blue}{\{data list\}}

Your mission is to judge which data this language model generate fits the prompt most and which fits worst, and explain your judgment. 

The JSON object you output must contain the following keys:\\
- "reason": a string, the reason of your judgment.\\
- "best": a number, the index of the generated data that fits prompt the most (indice start from 0).\\
- "worst": a number, the index of the generated data that fits prompt the worst.\\

Your output must always be a JSON object only, do not explain yourself or output anything else.

\end{prompt}
With this prompt, we can obtain a best and worst data of the data list evaluated by GPT-4.
Then, we can get preference data pairs based on the best and worst data.

For the SFT of $\pi^{\text{Re}}_{\theta}$, we use GPT-4 to evaluate the quality of synthetic data from multiple aspects and produce the revised data for training signals.

\begin{prompt}[title={Prompt: Generating Revise Data}, label=prompt:revise]

A language model has been given a prompt: \textcolor{blue}{\{data prompt\}}

The output generated by the model is: \textcolor{blue}{\{data example\}}

Your task is to evaluate the generated output based on the following criteria:

1. Relevance: Assess whether the output directly addresses the task described in the prompt.\\
2. Completeness: Check if the output includes all necessary elements as specified in the prompt.\\
3. Accuracy: Verify if the output is factually correct and adheres to the guidelines provided in the prompt.

For each criterion, provide a brief explanation supporting your evaluation. Then, provide a revised version of the output.

Your response should be a valid JSON object containing the following keys:

- "reason": A string providing the reason for your judgment.\\
- "revision": A string with the revised version of the output based on your evaluation and the prompt.

Ensure your output is always a valid JSON object, formatted as a JSON string. Do not include any additional explanations or information.

\end{prompt}

\section{Data Examples}
\label{appendix: example}

\subsection{Topics}

In order to mitigate the hallucination and introduce more diversity to LLMs, we propose to sample multi-grained topics from ODP.
Some examples of the sampled raw topics are presented in Table~\ref{tab:topic_example}.
Some of these topics are wide categories (\eg, ``Arts''), which will make LLM generate more abstract data.
And some of these topics are detailed and specific, which may cause the synthetic data to include some noisy information.
Therefore, we propose to truncate the topics with depth more than four by discarding their middle information.
For example, for ``Arts/Movies/Titles/3/36\_Hours\_-\_1964/Cast\_and\_Crew'', we will only keep ``Arts/Movies/36\_Hours\_-\_1964/Cast\_and\_Crew''.
By this, we can keep its main category and some details without introducing too much noise.

\begin{table*}[h]
\centering
\begin{tabular}{l}
\toprule

Society/Crime/Criminals/Outlaws/Bonnie\_and\_Clyde \\
Sports/Baseball/People/Players/E/Estes,\_Shawn \\
Arts/Performing\_Arts/Dance/Folk/Square\_Dancing/Clubs/United\_States/Oregon \\
Regional/Europe/United\_Kingdom/England/County\_Durham/Darlington/Business\_and\_Economy/Shopping \\
Business/Food\_and\_Related\_Products/Produce/Frozen \\
Arts/Performing\_Arts/Acting/Actors\_and\_Actresses/V/Vaughn,\_Robert/Movies \\
Sports/Hockey/Ice\_Hockey/Players \\
Science/Biology/Flora\_and\_Fauna/Animalia/Arthropoda/Insecta/Diptera/Rhagionidae \\
Games/Video\_Games/Action/S/Snake\_Games/Downloads/Free \\
Regional/Asia/South\_Korea/Jeonnam/Yeonggwang \\
Computers/CAD\_and\_CAM/Electronic\_Design\_Automation \\
Regional/Europe/France/Regions/Languedoc-Roussillon/Lozere \\
Arts/Movies/Titles/3/36\_Hours\_-\_1964/Cast\_and\_Crew \\
Science/Technology/Structural\_Engineering/Bridge/History/People/Beedy,\_Daniel \\
Regional/Middle\_East/Cyprus/Limassol\_District/Travel\_and\_Tourism/Accommodation \\
Recreation/Food/Drink/Wine/Events/United\_States/Texas \\
Health/Medicine/Medical\_Specialties/Ophthalmology/Refractive\_Correction/LASIK \\
Arts \\
Society/Issues/Business/Allegedly\_Unethical\_Firms/Halliburton/Opposing\_Views \\

\bottomrule
\end{tabular}
\caption{Examples of topics sampled from ODP without truncation.}
\label{tab:topic_example}
\end{table*}

\subsection{Alignment Data}

\begin{figure*}[!t]
	\centering
	\includegraphics[width=0.95\textwidth]{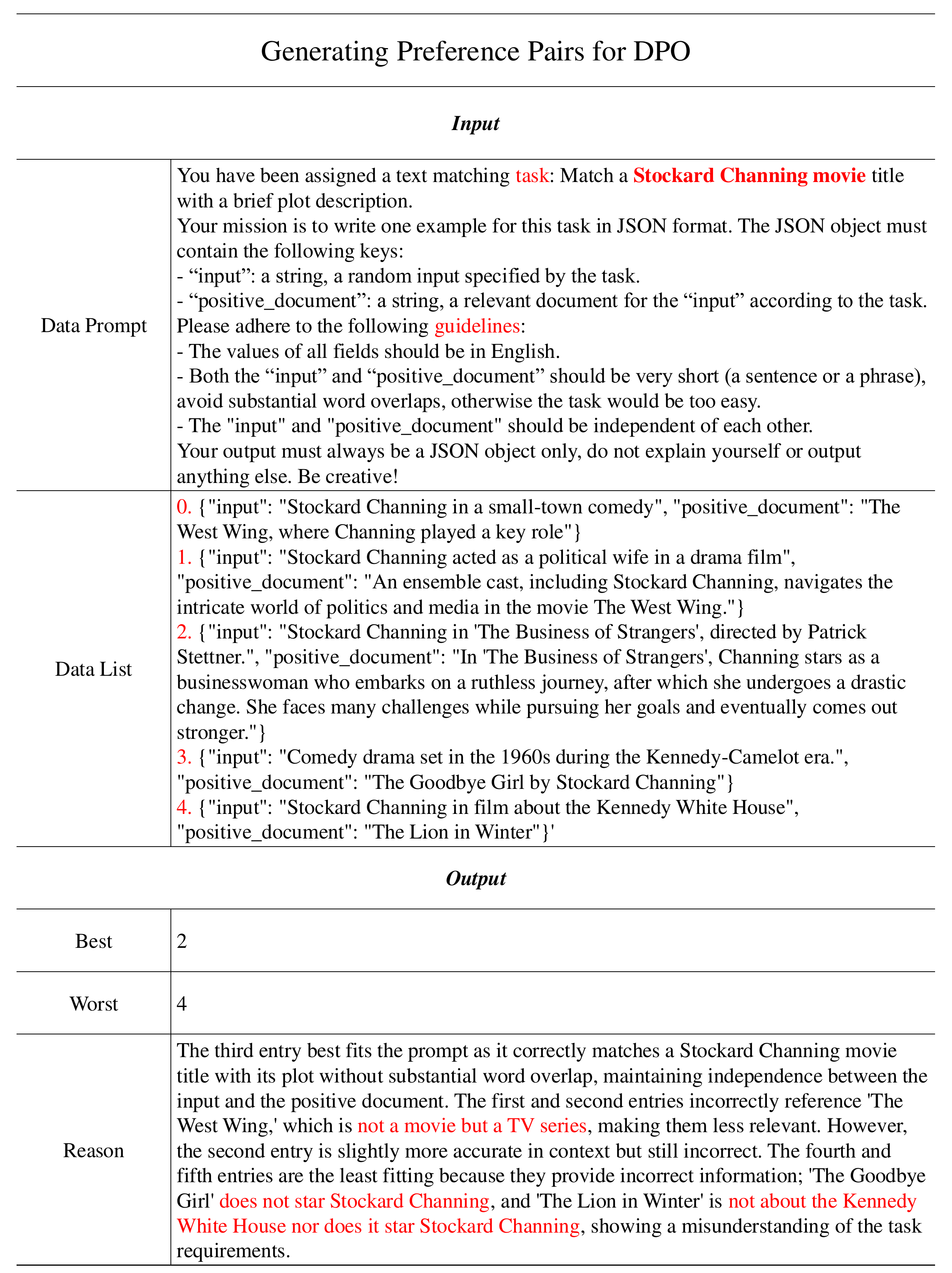}
	\caption{An example to show the generated preference signals for DPO. A data prompt and a data list are fed into GPT-4 and it evaluates the best and worst data according to the requirements of prompt. The data prompt template is from E5$_\text{mistral}$~\cite{E5mistral}.}
	\vspace{-2ex}
	\label{fig:dpo_example}
\end{figure*}

\begin{figure*}[!t]
	\centering
	\includegraphics[width=0.95\textwidth]{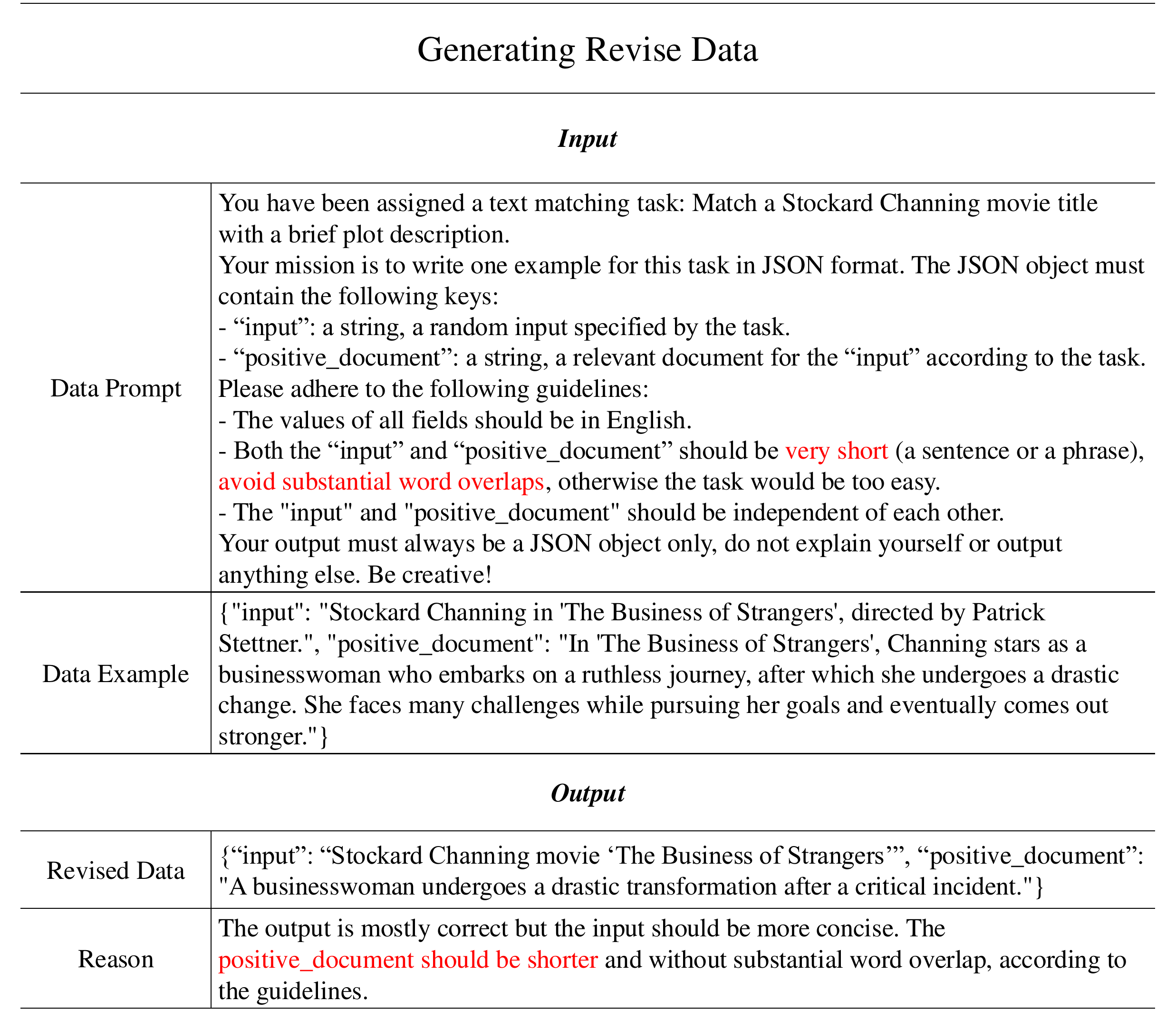}
	\caption{An example to show the generated revision signals for SFT the data revisor. A data prompt and a data list are fed into GPT-4 and it improves the data based on the given guidelines in the prompt. }
	\vspace{-2ex}
	\label{fig:revise_example}
\end{figure*}

In this section, we present data used for aligning $\pi^{\text{Sr}}_{\theta}$ and $\pi^{\text{Re}}_{\theta}$ in Figure~\ref{fig:dpo_example} and Figure~\ref{fig:revise_example}, respectively.

\subsection{Synthetic Embedding Data}

\begin{figure*}[!t]
	\centering
	\includegraphics[width=0.95\textwidth]{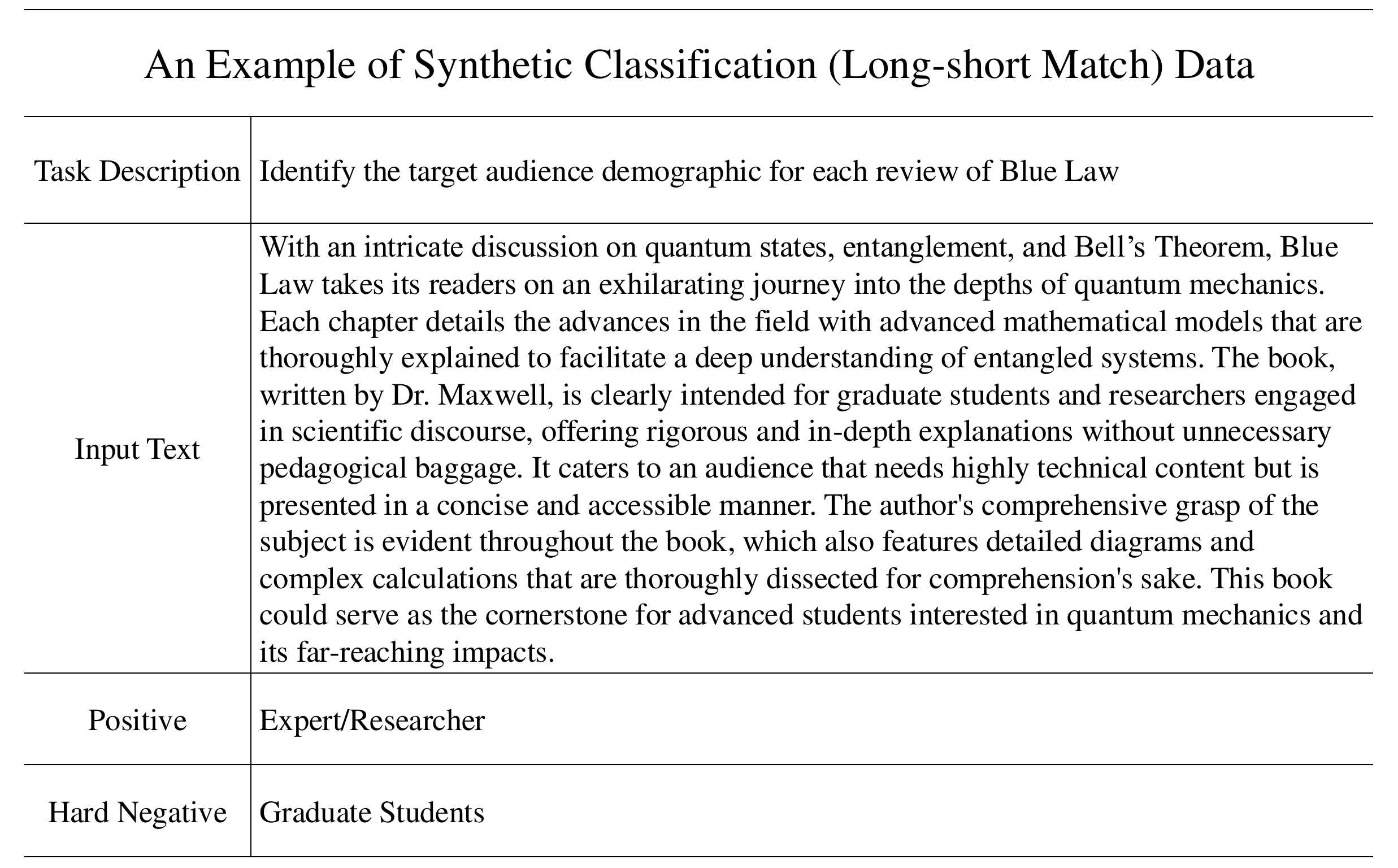}
	\caption{An example of the synthetic classification data. The data prompt template is from E5$_\text{mistral}$~\cite{E5mistral}.}
	\vspace{-2ex}
	\label{fig:cla_example}
\end{figure*}

\begin{figure*}[!t]
	\centering
	\includegraphics[width=0.95\textwidth]{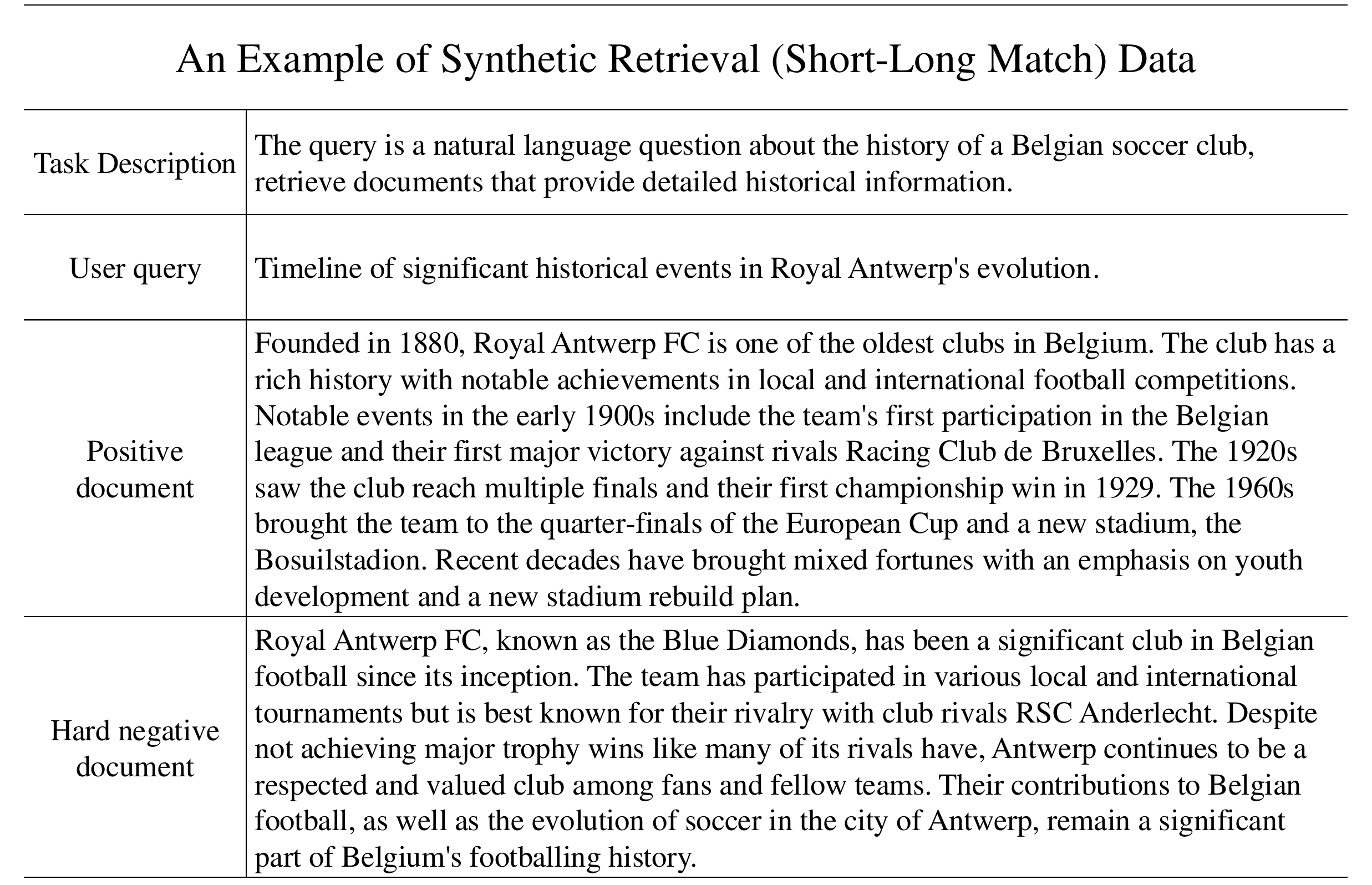}
	\caption{An example of the synthetic retrieval data. }
	\vspace{-2ex}
	\label{fig:ret_example}
\end{figure*}

\begin{figure*}[!t]
	\centering
	\includegraphics[width=0.95\textwidth]{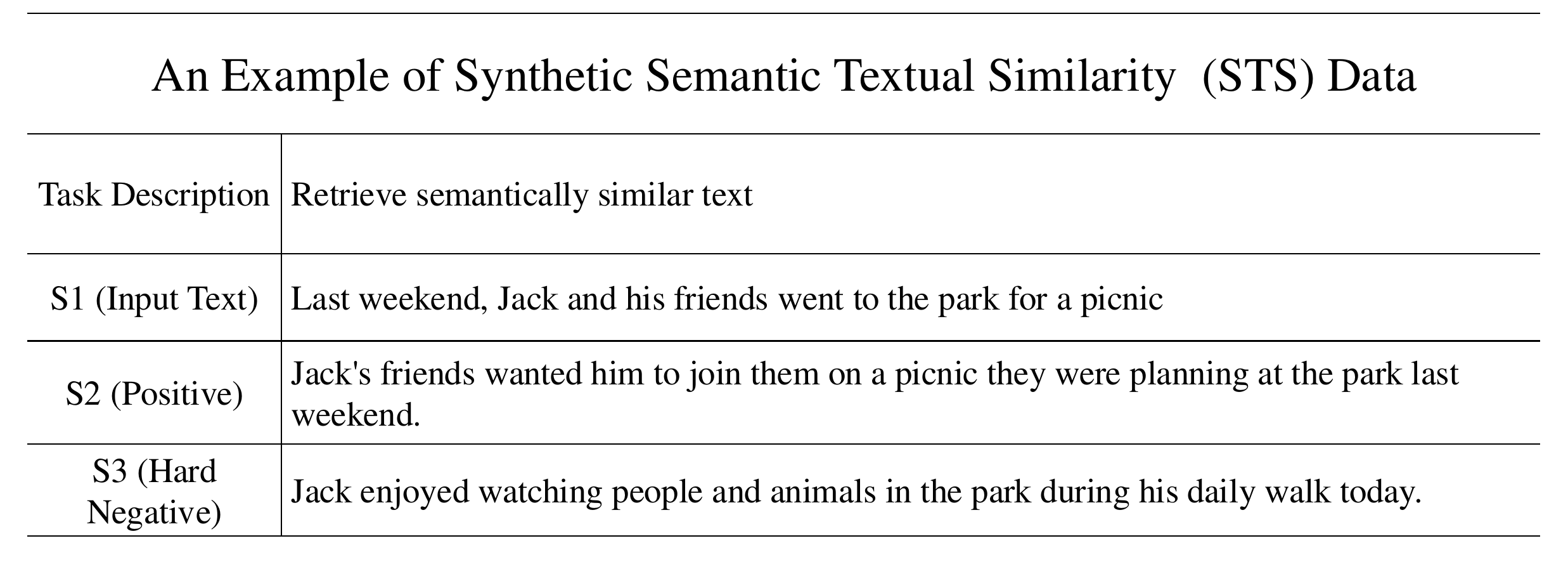}
	\caption{An example of the synthetic semantic textual similarity data. }
	\vspace{-2ex}
	\label{fig:sts_example}
\end{figure*}

\begin{figure*}[!t]
	\centering
	\includegraphics[width=0.95\textwidth]{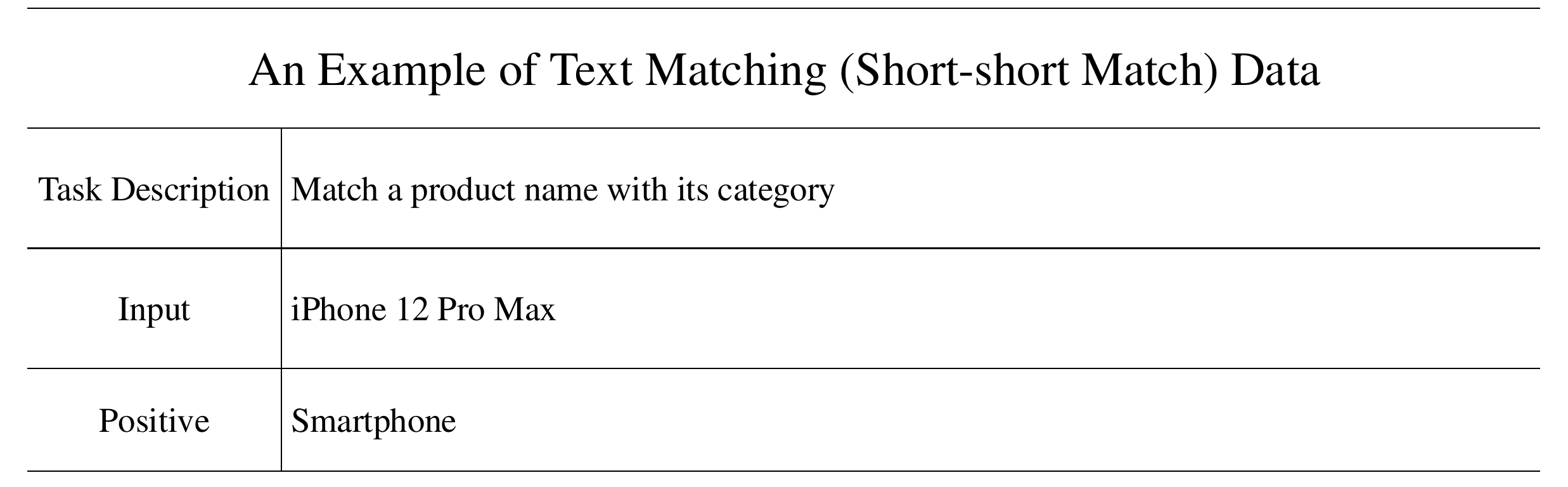}
	\caption{An example of the synthetic short-short matching data. }
	\vspace{-2ex}
	\label{fig:s2s_example}
\end{figure*}

\begin{figure*}[!t]
	\centering
	\includegraphics[width=0.95\textwidth]{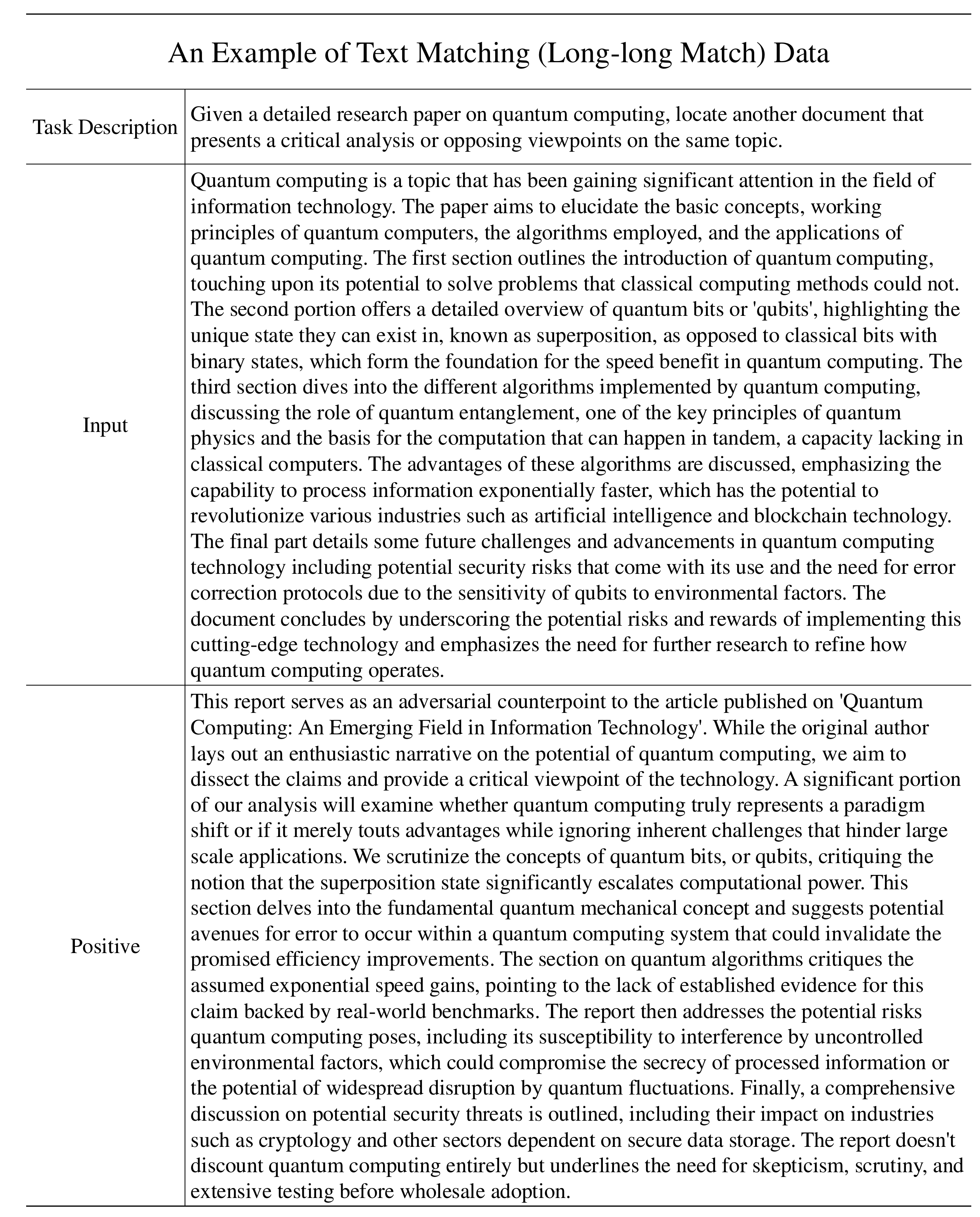}
	\caption{An example of the synthetic long-long matching data. }
	\vspace{-2ex}
	\label{fig:p2p_example}
\end{figure*}

In this section, we present examples of synthetic data of various task types in Figure~\ref{fig:cla_example} (classification), Figure~\ref{fig:ret_example} (retrieval), Figure~\ref{fig:sts_example} (STS), Figure~\ref{fig:s2s_example} (short-short matching), and Figure~\ref{fig:p2p_example} (long-long matching).

\section{Detailed Results}

In this section, we present detailed evaluation results of \ours{} in zero-shot setting and full-data setting.
The results on all 56 datasets of MTEB benchmark are shown in Table~\ref{tab:detailed_results}.

\begin{table*}[]
\small
\centering
\begin{tabular}{lcc}
\toprule 
Dataset                          & w/ synthetic only & full data     \\ \hline
BIOSSES                          & 85.4              & 87.1          \\
SICK-R                           & 79.6              & 82.5          \\
STS12                            & 77.7              & 80.2          \\
STS13                            & 87.9              & 89.9          \\
STS14                            & 81.8              & 86.2          \\
STS15                            & 87.7              & 91.2          \\
STS16                            & 85.8              & 88.2          \\
STS17                            & 86.4              & 91.9          \\
STS22                            & 69.2              & 68.3          \\
STSBenchmark                     & 84.7              & 89.2          \\
SummEval                         & 31.7              & 31.1          \\
SprintDuplicateQuestions         & 95.8              & 95.5          \\
TwitterSemEval2015               & 77.5              & 81.7          \\
TwitterURLCorpus                 & 85.7              & 87.4          \\
AmazonCounterfactualClassification       & 78.2              & 76.7          \\
AmazonPolarityClassification     & 95.7              & 96.2          \\
AmazonReviewsClassification      & 56.7              & 56.3          \\
Banking77Classification          & 87.7              & 88.6          \\
EmotionClassification            & 52.3              & 51.0          \\
ImdbClassification               & 93.9              & 94.9          \\
MassiveIntentClassification      & 79.0              & 80.2          \\
MassiveScenarioClassification    & 81.5              & 82.3          \\
MTOPDomainClassification         & 95.4              & 95.9          \\
MTOPIntentClassification         & 86.0              & 87.1          \\
ToxicConversationsClassification & 68.9              & 68.4          \\
TweetSentimentExtractionClassification   & 64.3              & 63.8          \\
AskUbuntuDupQuestions            & 65.5              & 67.2          \\
MindSmallReranking               & 33.0              & 33.4          \\
SciDocsRR                        & 86.7              & 87.3          \\
StackOverflowDupQuestions        & 53.9              & 55.2          \\
ArxivClusteringP2P               & 50.7              & 51.1          \\
ArxivClusteringS2S               & 46.4              & 47.0          \\
BiorxivClusteringP2P             & 42.6              & 42.0          \\
BiorxivClusteringS2S             & 39.7              & 39.6          \\
MedrxivClusteringP2P             & 35.1              & 37.0          \\
MedrxivClusteringS2S             & 35.7              & 36.3          \\
RedditClustering                 & 56.1              & 57.9          \\
RedditClusteringP2P              & 63.9              & 65.3          \\
StackExchangeClustering          & 70.0              & 71.6          \\
StackExchangeClusteringP2P       & 39.9              & 39.0          \\
TwentyNewsgroupsClustering       & 54.9              & 55.2          \\
ArguAna                          & 40.5              & 59.3          \\
ClimateFEVER                     & 22.2              & 37.8          \\
CQADupstackAndroidRetrieval      & 41.7              & 41.6          \\
DBPedia                          & 43.3              & 49.7          \\
FEVER                            & 77.7              & 88.5          \\
FiQA2018                         & 39.5              & 56.1          \\
HotpotQA                         & 55.6              & 75.2          \\
MSMARCO                          & 25.9              & 42.5          \\
NFCorpus                         & 36.5              & 38.7          \\
NQ                               & 53.3              & 61.7          \\
QuoraRetrieval                   & 84.6              & 89.3          \\
SCIDOCS                          & 21.0              & 16.6          \\
SciFact                          & 71.9              & 77.2          \\
Touche2020                       & 23.8              & 25.8          \\
TRECCOVID                        & 83.9              & 87.4          \\ \hline
Average                          & 63.4              & \textbf{66.5} \\ \bottomrule
\end{tabular}
\caption{Detailed results of \ours{} in the zero-shot setting and full-data setting on each dataset of MTEB. 
The details about the evaluation metrics and dataset statistics can be found in its original paper~\cite{mteb}}
\label{tab:detailed_results}
\end{table*}

\end{document}